\documentclass[11pt]{article}

\usepackage[]{acl}

\usepackage{times}
\usepackage{latexsym}

\usepackage[T1]{fontenc}
\usepackage[utf8]{inputenc}
\usepackage{microtype}
\usepackage{inconsolata}
\usepackage{booktabs}
\usepackage{multirow}
\usepackage{colortbl}
\definecolor{oursgreen}{RGB}{209,231,211}
\definecolor{raddblue}{RGB}{44,100,166}
\definecolor{raddgreen}{RGB}{33,133,88}
\definecolor{raddorange}{RGB}{217,115,37}
\usepackage{makecell}
\usepackage{amsmath}
\usepackage{amssymb}
\usepackage{graphicx}
\usepackage{algorithm}
\usepackage{algpseudocode}
\usepackage{tikz}
\usepackage{pgfplots}
\pgfplotsset{compat=1.18}
\usepgfplotslibrary{groupplots}
\usepackage{placeins}
\usetikzlibrary{arrows.meta,positioning,shapes.geometric,fit,backgrounds,calc,decorations.pathreplacing}

\title{RADD: Retrieval-Anchored Discrete Diffusion for Multi-Modal\\Knowledge Graph Completion}

\author{Guanglin Niu \and Bo Li \\
        School of Artificial Intelligence, Beihang University \\
        \texttt{\{beihangngl, boli\}@buaa.edu.cn}}

\begin{document}

\maketitle

\begin{abstract}
Multi-modal knowledge graph completion aims to infer missing links by integrating structural, visual, and textual information. However, global retrieval can leave fine-grained entity ambiguity unresolved, while shortlist-only prediction excludes answers missed by candidate selection. We propose \textbf{R}etrieval-\textbf{A}nchored \textbf{D}iscrete \textbf{D}iffusion (RADD), a unified framework that turns retrieval uncertainty into a probabilistic reference for multimodal refinement. By anchoring discrete diffusion to the query-specific retrieval distribution, RADD learns how complementary multimodal evidence should revise entity preferences, coupling global retrieval with fine-grained disambiguation. Bidirectional learning unifies head and tail refinement, while coverage-aware inference adapts ranking corrections to candidate coverage without restricting predictions to the retrieved set. Experiments on three MMKGC benchmarks demonstrate RADD's effectiveness and consistent improvements over strong baselines and its matched retriever.
\end{abstract}

\section{Introduction}
\label{sec:introduction}

Multi-modal knowledge graph completion (MMKGC) predicts missing entities from structural, visual, and textual evidence~\cite{liu2019mmkg,wang2023tivakg,chen2024kgmultimodalsurvey}. Knowledge graph embedding (KGE) models support global scoring~\cite{bordes2013transe,trouillon2016complex,sun2019rotate}, enhanced by relation-aware fusion~\cite{lee2023vista,zhang2024native,zhang2025momok} and language-model enrichment~\cite{guo2024mkgl,guo2025kon}. Yet strong retrieval still confuses entities sharing neighborhoods or multimodal content, motivating fine-grained candidate comparison.

\begin{figure}[t]
\centering
\includegraphics[width=\columnwidth]{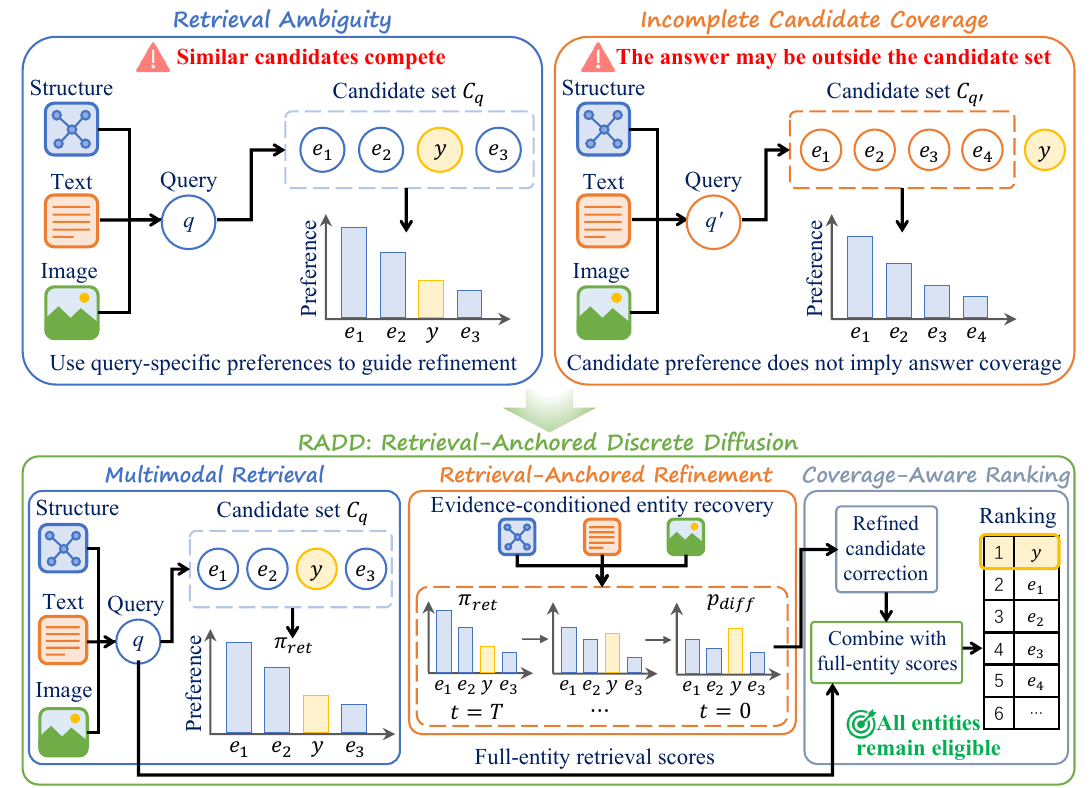}
\caption{RADD addresses retrieval ambiguity and incomplete candidate coverage (top). Multimodal evidence refines retrieval preferences $\pi_{\mathrm{ret}}$ into $p_{\mathrm{diff}}$, and coverage-aware corrections update full-entity scores, keeping all entities eligible (bottom).}
\label{fig:introduction}
\end{figure}

Our central insight is that retrieval uncertainty should be the starting point of refinement, not merely a constraint on its output. A query-specific retrieval distribution expresses which entities are plausible and how strongly they compete. As shown in Fig.~\ref{fig:introduction}, using this distribution as a reference lets refinement focus on complementary evidence that revises candidate preferences. Realizing this idea requires integrating retrieval preferences into the refinement process while accounting for the possibility that the candidate set does not contain the correct answer.

To turn this retrieval reference into evidence-driven refinement, discrete diffusion provides a probabilistic mechanism for revising preferences over entity identities~\cite{austin2021structured}. However, a query-independent diffusion endpoint does not retain the retriever's query-specific preferences. We therefore anchor diffusion to the retrieval distribution, so that multimodal denoising learns how evidence should revise these preferences.

Refining candidate preferences alone, however, does not address the possibility that the correct answer is absent from the candidate set. A distribution normalized over candidates represents relative preferences within the set, but does not explicitly model answer inclusion. Refinement must therefore account for candidate coverage while preserving full-entity ranking, allowing local evidence to influence predictions without excluding candidate-external answers.

To address both requirements, we propose \textbf{R}etrieval-\textbf{A}nchored \textbf{D}iscrete \textbf{D}iffusion (RADD), a unified framework that couples global retrieval with evidence-driven candidate refinement. RADD uses a query-specific retrieval distribution as both the diffusion endpoint and the reference for evidence-conditioned recovery. The denoiser uses structural and multimodal evidence to revise relative candidate preferences, while full-state reverse propagation integrates competing recovery paths without trajectory sampling. Bidirectional training jointly learns entity recovery and candidate coverage from cross-fitted head and tail queries with natural candidate sets. Predicted coverage then moderates a density-ratio correction to full-entity scores, adapting refinement to answer inclusion while keeping all entities eligible for prediction. Experiments on DB15K, MKG-W, and MKG-Y show that RADD outperforms the strongest published baselines, demonstrating the effectiveness of retrieval-anchored refinement.

Our contributions are:
\begin{itemize}
\item We introduce retrieval-anchored discrete diffusion, coupling a query-specific corruption endpoint with reference-relative entity recovery. Full-state reverse propagation integrates candidate alternatives to refine retrieval preferences using complementary evidence.
\item We develop coverage-aware, bidirectional refinement on natural candidates. A learned inclusion probability moderates density-ratio updates, enabling candidate refinement within full-entity ranking and recovering retrieval when correction is inactive.
\item Experiments on three benchmarks show leading reported performance and consistent gains beyond a strong matched retriever. Ablations, directional analyses, and sensitivity studies substantiate its effectiveness across prediction directions and candidate budgets.
\end{itemize}

\section{Related Work}

\paragraph{KGE-based KGC.}
Translation-based~\cite{bordes2013transe}, bilinear~\cite{yang2015distmult,trouillon2016complex,balazevic2019tucker}, and rotational~\cite{sun2019rotate,chao2021pairre,tang2020gcote} KGE models score triples using entity and relation embeddings. Graph neural-network~\cite{zhu2021nbfnet} and language-model scorers~\cite{yao2019kgbert} extend the available scoring mechanisms. MMKGC adds uneven modality coverage and relation-dependent feature importance~\cite{wang2017kgesurvey,chen2024kgmultimodalsurvey}, motivating query-specific evidence use.

\paragraph{Multi-modal fusion for KGC.}
Early MMKGC models inject visual or textual features directly into KGE objectives~\cite{xie2017ikrl,sergieh2018tbkgc,pezeshkpour2018emrd,wang2019transae,lu2022mmkrl}. Subsequent work improves fusion quality via optimal transport~\cite{cao2022otkge}, cross-modal attention~\cite{li2023imf}, hybrid transformers~\cite{lee2023vista}, modality-aware negative sampling~\cite{xu2022mmrns,zhang2023mans}, adaptive weighting under incomplete observations~\cite{zhang2024adamfmat,zhang2024native}, and relation-conditioned modality routing~\cite{zhang2025momok}. LLM-based methods further adapt language models to entity-level prediction~\cite{guo2024mkgl,guo2025kon}. Unlike ReranKGC~\cite{gao2025rerankgc}, which combines an embedding top-$K$ retriever, relation-specific prompt-tuned KGC-CLIP, and a relation-wise ensemble, RADD learns retrieval-anchored entity recovery with evidence-conditioned density ratios and coverage-moderated full-entity ranking.

\paragraph{Diffusion models for structured prediction.}
Continuous DDPM~\cite{ho2020ddpm} and discrete D3PM~\cite{austin2021structured} establish denoising-based generation in continuous and categorical spaces. D3PM supports structured transition kernels beyond uniform or absorbing noise. DiffusionCom~\cite{huang2025diffusioncom} applies Gaussian diffusion to an entity-probability representation for MMKGC. RADD instead operates on discrete entity identities and couples a retrieval-defined endpoint with relative-evidence denoising and coverage-aware ranking. Its contribution is this retrieval-centered formulation, building on the general categorical-diffusion framework.

\begin{figure*}[t]
\centering
\includegraphics[width=\textwidth]{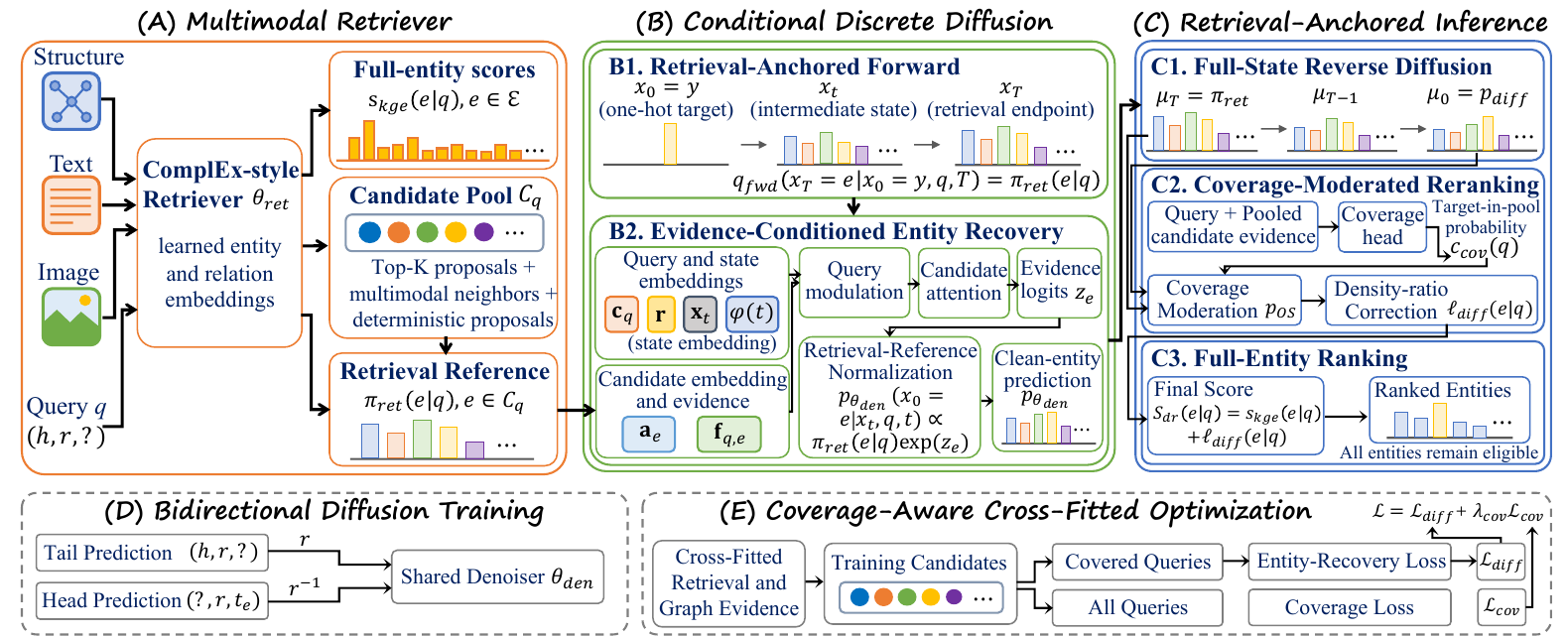}
\caption{Overview of RADD. (A) Retrieval supplies scores, candidates, and $\pi_{\mathrm{ret}}$. (B) Retrieval-anchored corruption trains evidence-conditioned recovery. (C) Reverse diffusion yields coverage-moderated global score updates. (D, E) Bidirectional cross-fitted training supervises recovery on covered queries and coverage on all queries.}
\label{fig:overview}
\end{figure*}

\section{Task Definition}

Let $\mathcal{G}=(\mathcal{E},\mathcal{R},\mathcal{T})$ be a multi-modal KG with entity set $\mathcal E$, relation set $\mathcal R$, and observed training triples $\mathcal T\subseteq\mathcal E\times\mathcal R\times\mathcal E$. Entities have structural context and visual/textual features. In $(h,r,t_e)$, $h,t_e\in\mathcal E$ are head/tail entities and $r\in\mathcal R$ is their relation. A query $q=(h,r,?)$ or $q=(?,r,t_e)$ replaces the unknown entity by $?$; $y$ denotes its correct answer. RADD refines natural candidates while ranking all entities.

\section{Approach}
\label{sec:approach}

As illustrated in Fig.~\ref{fig:overview}, RADD formulates candidate refinement as evidence-conditioned entity recovery through retrieval-anchored discrete diffusion. The retriever supplies full-entity scores and a query-specific diffusion endpoint. The evidence-conditioned denoiser predicts clean-entity probabilities relative to the retrieval reference, and these predictions parameterize the reverse transitions for diffusion-based refinement. Predicted coverage moderates the ranking update. We describe retrieval and query representations, conditional diffusion, bidirectional and coverage-aware training, and full-entity inference.

\subsection{Multimodal Retrieval and Query~Representation}
\label{sec:backbone}

A pretrained ComplEx-based retriever~\cite{trouillon2016complex} learns complex-valued entity/relation embeddings and returns full-entity scores $s_{\mathrm{kge}}(e\mid q)$. These scores select candidates and underpin final ranking; their standardized version defines the retrieval reference. Retrieval parameters $\theta_{\mathrm{ret}}$ remain frozen during refinement training and fixed across matched controls.

For tail prediction, the observed entity is $o(q)=h$ and the directed relation is $\rho(q)=r$; for head prediction, $o(q)=t_e$ and $\rho(q)=r^{-1}$. Refinement uses separate trainable lookup tables: $\mathbf a_e$ denotes the embedding of entity $e$, and $\mathbf r$ that of $\rho(q)$, including inverse relations. The query context is $\mathbf c_q=\mathbf a_{o(q)}$. These embeddings are jointly optimized with the shared denoiser and coverage head as $\theta_{\mathrm{den}}$, independently of the frozen $\theta_{\mathrm{ret}}$.

\subsection{Conditional Discrete Diffusion}
\label{sec:diffusion}

Conditional discrete diffusion refines a distribution over candidate entity identities. Let $x_t$ be the categorical entity state at diffusion step $t\in\{0,\ldots,T\}$, where $T$ is the number of steps; $t_e$ continues to denote a tail entity. Natural candidates are constructed without inserting the target, preserving retrieval ambiguities and misses during training.

For proposal width $K$, $\operatorname{TopK}$ returns the set of $K$ highest-scoring entity IDs:
\begin{equation}
\mathcal{C}_K(q) = \operatorname{TopK}_{e \in \mathcal{E}}\, s_{\mathrm{kge}}(e \mid q).
\label{eq:shortlist}
\end{equation}
We expand this initial set $\mathcal C_K(q)$ with visual/textual neighbors and deterministic proposals to obtain $\mathcal C_q$, of size $K_q=|\mathcal C_q|$ (Appendix~\ref{app:algorithm}). For candidate $e\in\mathcal C_q$, retrieval defines
\begin{equation}
\pi_{\mathrm{ret}}(e\mid q)=\frac{\exp b_{q,e}}{\sum_{e'\in\mathcal C_q}\exp b_{q,e'}},
\label{eq:retriever_reference}
\end{equation}
Here $b_{q,e}=\mathrm Z_{\mathcal E}(s_{\mathrm{kge}}(e\mid q))$. The operator $\mathrm Z_{\mathcal E}$ subtracts the query-wise mean and divides by the stabilized standard deviation over $\mathcal E$. Thus $\pi_{\mathrm{ret}}$ is positive on $\mathcal C_q$, not a full-entity posterior.

\paragraph{Retrieval-anchored forward process.}
At each step, the process retains the current identity with probability $1-\beta_t$ and otherwise draws from the retrieval reference:
\begin{equation}
Q_t^q(j\mid i)=(1-\beta_t)\mathbb I[j=i]+\beta_t\pi_{\mathrm{ret}}(j\mid q),
\label{eq:transition}
\end{equation}
Here $Q_t^q(j\mid i)$ is the transition probability from $x_{t-1}=i$ to $x_t=j$, with $i,j\in\mathcal C_q$. The indicator $\mathbb I[\cdot]$ is 1 when its condition holds and 0 otherwise; $0<\beta_t\leq1$ is the replacement probability. Per-step and cumulative retention probabilities are $\alpha_t=1-\beta_t$ and $\bar\alpha_t=\prod_{u=1}^t\alpha_u$. For clean target $x_0=y\in\mathcal C_q$, the exact forward marginal~\cite{austin2021structured}, $\bar Q_t^q(j\mid y)=q_{\mathrm{fwd}}(x_t=j\mid x_0=y,q,t)$, is
\begin{equation}
\begin{aligned}
\bar Q_t^q(j\mid y)={}&\bar\alpha_t\mathbb I[j=y]\\
&+(1-\bar\alpha_t)\pi_{\mathrm{ret}}(j\mid q).
\end{aligned}
\label{eq:forward}
\end{equation}
We use a cosine schedule with $\bar\alpha_0=1$ and $\bar\alpha_T=0$, enforcing $\beta_T=1$. Thus $q_{\mathrm{fwd}}(x_T\mid x_0,q,T)=\pi_{\mathrm{ret}}$: training ends at exactly the reference available to initialize inference. The closed-form marginal also permits direct sampling of any intermediate state without simulating preceding transitions.

\paragraph{Evidence-conditioned entity recovery.}
With state embedding $\mathbf x_t=\mathbf a_{x_t}$ and learned timestep embedding $\phi(t)$, the denoiser input is
\begin{equation}
\tilde{\mathbf{x}}(q,t,x_t) = \bigl[\mathbf{c}_q;\; \mathbf{r};\; \mathbf x_t;\; \phi(t)\bigr],
\label{eq:denoiser_input}
\end{equation}
where $[\cdot;\cdot]$ denotes concatenation. Candidate entity embeddings $\mathbf a_e$ and 24-dimensional structural/multimodal descriptors $\mathbf f_{q,e}$ undergo query-modulated attention, yielding $\mathbf H_{q,t}\in\mathbb R^{K_q\times d_{\mathrm{den}}}$ with hidden width $d_{\mathrm{den}}$ (Appendix~\ref{app:algorithm}). This matrix is shared across current states at fixed $(q,t)$. The network returns $\mathbf z\in\mathbb R^{K_q}$. With $z_e=[\mathbf z(q,t,x_t)]_e$ and $\pi_e=\pi_{\mathrm{ret}}(e\mid q)$, we obtain
\begin{equation}
\begin{gathered}
\mathbf z(q,t,x_t)=f_{\theta_{\mathrm{den}}}(\tilde{\mathbf x}(q,t,x_t),\mathbf H_{q,t}),\\
p_{\theta_{\mathrm{den}}}(x_0=e\mid x_t,q,t)
=\frac{\pi_e\exp z_e}{\sum_{e'\in\mathcal C_q}\pi_{e'}\exp z_{e'}}.
\end{gathered}
\label{eq:reverse}
\end{equation}
Thus $f_{\theta_{\mathrm{den}}}$ returns logits whose exponentials reweight $\pi_{\mathrm{ret}}$; the normalized $p_{\theta_{\mathrm{den}}}$ predicts clean identity $x_0$ for reverse transitions, not final ranking. Candidate-independent logits recover $\pi_{\mathrm{ret}}$. Raw retrieval scores/ranks are excluded from neural inputs; retrieval enters via the reference and final~ranking.

\subsection{Bidirectional Diffusion Training}

Head and tail queries provide complementary recovery tasks for the shared denoiser, distinguished by their directed relations. Let the binary label $z_q=\mathbb I[y\in\mathcal C_q]$ indicate target coverage, distinct from the evidence logits $z_e$. For covered queries, we sample $t$ uniformly from $\{1,\ldots,T\}$ and $x_t$ from Eq.~\ref{eq:forward}, giving
\begin{equation}
\begin{aligned}
\mathcal L_{\mathrm{diff}}={}&\mathbb E_{(q,y):z_q=1}\mathbb E_{t,x_t}\\
&[-\log p_{\theta_{\mathrm{den}}}(x_0=y\mid x_t,q,t)].
\end{aligned}
\label{eq:diff_loss}
\end{equation}
Here $\mathbb E$ denotes expectation over covered training pairs and the sampled $(t,x_t)$. This recovery cross-entropy uses both directions; uncovered queries supervise only coverage, without target insertion.

\subsection{Coverage-Aware Cross-Fitted~Optimization}

To expose refinement to realistic retrieval errors, we split training triples into three folds, keeping both query directions together. Each fold receives scores, candidates, and graph-derived evidence constructed without its facts. These cross-fitted inputs remain fixed during denoiser optimization; evaluation uses retrieval trained on the full training set.

The coverage head predicts inclusion in the expanded pool: $c_{\mathrm{cov}}(q)=p_{\theta_{\mathrm{den}}}(z_q=1\mid q,\mathcal C_q)\in[0,1]$. It uses query context and attention-pooled candidates, not target $y$. Target membership supervises binary cross-entropy over all training pairs, covered or not:
\begin{equation}
\begin{aligned}
\mathcal L_{\mathrm{cov}}=\mathbb E_{(q,y)}[{}&-z_q\log c_{\mathrm{cov}}(q)\\
&-(1-z_q)\log(1-c_{\mathrm{cov}}(q))].
\end{aligned}
\label{eq:coverage_loss}
\end{equation}
The denoiser-stage objective is
\begin{equation}
\mathcal L=\mathcal L_{\mathrm{diff}}+\lambda_{\mathrm{cov}}\mathcal L_{\mathrm{cov}},
\label{eq:total_loss}
\end{equation}
where $\lambda_{\mathrm{cov}}\geq0$ weights coverage supervision. Only $\theta_{\mathrm{den}}$ is optimized: recovery learns candidate preferences, while coverage learns target inclusion. Retrieval enters diffusion through the fixed reference, without a separate distillation loss.

\subsection{Retrieval-Anchored Inference}

\paragraph{Full-state reverse diffusion.}
Let $\mu_t$ be the column vector of state probabilities $\mu_t(i)$ over $i\in\mathcal C_q$, with query dependence suppressed. Initialize $\mu_T=\pi_{\mathrm{ret}}$. For current state $j$, the clean prediction weights Bayes posteriors to define the reverse transition to previous state $i$:
\begin{equation}
\begin{aligned}
R_t^q(i\mid j)={}&\sum_{e\in\mathcal C_q}p_{\theta_{\mathrm{den}}}(x_0=e\mid x_t=j,q,t)\\
&\quad\cdot\frac{\bar Q_{t-1}^q(i\mid e)Q_t^q(j\mid i)}{\bar Q_t^q(j\mid e)}.
\end{aligned}
\label{eq:learned_reverse}
\end{equation}
Here $i,j,e\in\mathcal C_q$ index previous, current, and hypothesized clean entities. The fraction is the reverse conditional given $x_0=e$ (Appendix~\ref{app:algorithm}); the sum marginalizes $e$, never the test answer. Probability mass propagates over all current states:
\begin{equation}
\mu_{t-1}(i)=\sum_{j\in\mathcal C_q}\mu_t(j)R_t^q(i\mid j).
\label{eq:reverse_propagation}
\end{equation}
Repeating for $t=T,\ldots,1$ yields $p_{\mathrm{diff}}(e\mid q)=\mu_0(e)$, distinct from a single clean-entity prediction. This exactly marginalizes the learned chain without trajectory sampling; the learned transitions need not equal the true reverse conditionals. Appendix~\ref{app:algorithm} derives the kernel and its quadratic evaluation.

\paragraph{Coverage-moderated density-ratio correction.}
\textbf{Reranking} integrates refinement into global ranking. Predicted coverage defines the moderated candidate distribution, for $e\in\mathcal C_q$,
\begin{equation}
\begin{aligned}
p_{\mathrm{OS}}(e\mid q)={}&c_{\mathrm{cov}}(q)p_{\mathrm{diff}}(e\mid q)\\
&+(1-c_{\mathrm{cov}}(q))\pi_{\mathrm{ret}}(e\mid q).
\end{aligned}
\label{eq:coverage_mixture}
\end{equation}
Here OS denotes open-set awareness: $p_{\mathrm{OS}}$ is a candidate distribution, not a full-entity posterior. Lower coverage shrinks refinement toward retrieval; it predicts inclusion, not correction success. Final scores are
\begin{equation}
\begin{aligned}
\ell_{\mathrm{diff}}(e\mid q)&=\begin{cases}
\displaystyle\log\frac{p_{\mathrm{OS}}(e\mid q)}{\pi_{\mathrm{ret}}(e\mid q)}, & e\in\mathcal C_q,\\
0, & e\notin\mathcal C_q,
\end{cases}
\\
S_{\mathrm{dr}}(e\mid q)&=s_{\mathrm{kge}}(e\mid q)+\ell_{\mathrm{diff}}(e\mid q).
\end{aligned}
\label{eq:diff_rerank}
\end{equation}
Here $\ell_{\mathrm{diff}}$ is the additive log-density-ratio correction and $S_{\mathrm{dr}}$ the final full-entity score. Candidate-external scores stay unchanged, but their ranks can change. If $c_{\mathrm{cov}}=0$ or $p_{\mathrm{diff}}=\pi_{\mathrm{ret}}$, ranking exactly recovers retrieval. The deployed score scale is retained without an interpolation parameter (Appendix~\ref{app:algorithm}).

\section{Experiments}

\subsection{Datasets}

\begin{table}[t]
\centering
\resizebox{\columnwidth}{!}{%
\begin{tabular}{lrrrrrrr}
\toprule
Dataset & \#Ent & \#Rel & Train & Valid & Test & Img & Txt \\
\midrule
DB15K & 12,842 & 279 & 79,222 & 9,904 & 9,902 & 12,818 & 9,078 \\
MKG-W & 15,000 & 169 & 34,196 & 4,276 & 4,274 & 14,463 & 14,123 \\
MKG-Y & 15,000 & 28 & 21,310 & 2,665 & 2,663 & 14,244 & 12,305 \\
\bottomrule
\end{tabular}}
\caption{Dataset statistics. Ent/Rel count entities/relations; Train/Valid/Test count triples. Img/Txt count entities with visual/textual features.}
\label{tab:data}
\end{table}

We evaluate on DB15K, MKG-W, and MKG-Y~\cite{zhang2025momok}, using the released splits and pre-extracted image/text features in Table~\ref{tab:data}.\footnote{DB15K follows MoMoK's released \href{https://github.com/zjukg/MoMoK/blob/master/datasets/DB15K/valid.txt}{validation} (9,904) and \href{https://github.com/zjukg/MoMoK/blob/master/datasets/DB15K/test.txt}{test} (9,902) files.} Each test triple generates one head and one tail query, giving 19,804/8,548/5,326 queries, respectively.

\subsection{Implementation Details}

\textbf{Retrieval and candidates:} We use a frozen ComplEx-based retriever (Appendix~\ref{app:retrieval-config}). We expand its top-$K$ proposals with visual/textual neighbors and deterministic proposals. Inference uses $K=256$ on DB15K and $K=64$ on MKG-W/Y, giving deduplicated pools of at most 288 and 96 entities, respectively. No target is inserted during training or inference.

\textbf{Training and selection:} Entity embedding dimension is 16 on DB15K and 32 on MKG-W/Y; hidden width $d_{\mathrm{den}}=96$, $T=4$, batch size 256, learning rate $\eta=3\times10^{-4}$, and $\lambda_{\mathrm{cov}}=1$. Training combines entity-recovery and coverage cross-entropy with seed 20260929. Three-fold cross-fitting constructs retrieval inputs; checkpoint selection is a separate step, using a training-derived holdout on DB15K and official validation sets on MKG-W/Y. The reported full-model checkpoint is epoch 2.

\textbf{Evaluation:} We rank over all entities and filter other known correct answers; candidate misses remain in evaluation. Mean reciprocal rank (MRR) averages reciprocal filtered ranks; Hits@$k$ (H@$k$, $k=1,3,10$) measures the fraction ranked within cutoff $k$. Metrics are percentages, micro-averaged over all head/tail queries; higher is better.

\begin{table*}[t]
\centering
\footnotesize
\setlength{\tabcolsep}{4.5pt}
\resizebox{\textwidth}{!}{%
\begin{tabular}{clcccccccccccc}
\toprule
\multicolumn{2}{c}{\multirow{2}{*}{Model}} & \multicolumn{4}{c}{DB15K} & \multicolumn{4}{c}{MKG-W} & \multicolumn{4}{c}{MKG-Y} \\
\cmidrule(lr){3-6}\cmidrule(lr){7-10}\cmidrule(lr){11-14}
\multicolumn{2}{c}{} & MRR & H@1 & H@3 & H@10 & MRR & H@1 & H@3 & H@10 & MRR & H@1 & H@3 & H@10 \\
\midrule
\multirow{7}{*}{\makecell{Unimodal\\KGC}}
& TransE & 24.86 & 12.78 & 31.48 & 47.07 & 29.19 & 21.06 & 33.20 & 44.23 & 30.73 & 23.45 & 35.18 & 43.37 \\
& TransD & 21.52 & 8.34 & 29.93 & 44.24 & 26.84 & 19.65 & 31.48 & 42.68 & 26.39 & 17.01 & 33.05 & 40.41 \\
& DistMult & 23.03 & 14.78 & 26.98 & 40.59 & 20.95 & 15.89 & 22.88 & 36.80 & 25.04 & 19.32 & 27.80 & 39.95 \\
& ComplEx & 27.48 & 18.13 & 31.57 & 45.37 & 28.09 & 21.45 & 30.89 & 44.77 & 28.94 & 23.11 & 31.07 & 43.48 \\
& RotatE & 29.28 & 17.87 & 36.12 & 49.66 & 33.67 & 26.80 & 36.68 & 46.73 & 34.95 & 29.10 & 38.35 & 45.30 \\
& PairRE & 31.13 & 21.62 & 35.91 & 49.30 & 34.40 & 28.24 & 36.71 & 46.04 & 32.01 & 25.53 & 35.83 & 43.89 \\
& GC-OTE & 31.85 & 22.11 & 36.52 & 51.18 & 33.92 & 26.55 & 35.96 & 46.05 & 32.95 & 26.77 & 36.44 & 44.08 \\
\midrule
\multirow{17}{*}{\makecell{MMKGC}}
& IKRL & 26.82 & 14.09 & 34.93 & 49.09 & 32.36 & 26.11 & 34.75 & 44.07 & 33.22 & 30.37 & 34.28 & 38.26 \\
& TBKGC & 28.40 & 15.61 & 37.03 & 49.86 & 31.48 & 25.31 & 33.98 & 43.24 & 33.99 & 30.47 & 35.27 & 40.07 \\
& TransAE & 28.09 & 21.25 & 31.17 & 41.17 & 30.00 & 21.23 & 34.91 & 44.72 & 28.10 & 25.31 & 29.10 & 33.03 \\
& MMKRL & 26.81 & 13.85 & 35.07 & 49.39 & 30.10 & 22.16 & 34.09 & 44.69 & 36.81 & 31.66 & 39.79 & 45.31 \\
& RSME & 29.76 & 24.15 & 32.12 & 40.29 & 29.23 & 23.36 & 31.97 & 40.43 & 34.44 & 31.78 & 36.07 & 39.09 \\
& VBKGC & 30.61 & 19.75 & 37.18 & 49.44 & 30.61 & 24.91 & 33.01 & 40.88 & 37.04 & 33.76 & 38.75 & 42.30 \\
& OTKGE & 23.86 & 18.45 & 25.89 & 34.23 & 34.36 & 28.85 & 36.25 & 44.88 & 35.51 & 31.97 & 37.18 & 41.38 \\
& MoSE & 28.38 & 21.56 & 30.91 & 41.67 & 33.34 & 27.78 & 33.94 & 41.06 & 36.28 & 33.64 & 37.47 & 40.81 \\
& IMF & 32.25 & 24.20 & 36.00 & 48.19 & 34.50 & 28.77 & 36.62 & 45.44 & 35.79 & 32.95 & 37.14 & 40.63 \\
& QEB & 28.18 & 14.82 & 36.67 & 51.55 & 32.38 & 25.47 & 35.06 & 45.32 & 34.37 & 29.49 & 36.95 & 42.32 \\
& VISTA & 30.42 & 22.49 & 33.56 & 45.94 & 32.91 & 26.12 & 35.38 & 45.61 & 30.45 & 24.87 & 32.39 & 41.53 \\
& AdaMF & 32.51 & 21.31 & 39.67 & 51.68 & 34.27 & 27.21 & 37.86 & 47.21 & 38.06 & 33.49 & 40.44 & 45.48 \\
& AdaMF-MAT & 35.14 & 25.30 & 41.11 & 52.92 & 35.85 & 29.04 & 39.01 & 48.42 & 38.57 & 34.34 & 40.59 & 45.76 \\
& MoMoK & 39.57 & \underline{32.38} & 43.45 & 54.14 & 35.89 & 30.38 & 37.54 & 46.13 & 37.91 & 35.09 & 39.20 & 43.20 \\
& SNAG & 36.30 & 27.40 & 41.10 & 53.00 & \underline{37.30} & 30.20 & \underline{40.50} & \underline{50.30} & \underline{39.50} & 35.40 & \underline{41.10} & \underline{47.10} \\
& MCKGC & \underline{39.79} & 31.92 & \underline{43.80} & \underline{54.66} & 36.88 & \underline{31.32} & 38.92 & 47.43 & 38.92 & \underline{35.49} & 40.57 & 45.21 \\
& ReranKGC & 37.80 & 30.50 & 41.50 & 52.60 & - & - & - & - & - & - & - & - \\
\midrule
\multirow{4}{*}{\makecell{Negative\\Sampling}}
& KBGAN & 24.83 & 19.48 & 27.15 & 36.49 & 25.78 & 19.05 & 28.71 & 37.89 & 24.26 & 20.68 & 25.68 & 29.16 \\
& MANS & 28.82 & 16.87 & 36.58 & 49.26 & 30.88 & 24.89 & 33.63 & 41.78 & 29.03 & 25.25 & 31.35 & 34.49 \\
& MMRNS & 32.68 & 23.01 & 37.86 & 51.01 & 35.03 & 28.59 & 37.49 & 47.47 & 35.93 & 30.53 & 39.07 & 45.47 \\
& DHNS & 34.59 & 26.80 & 39.20 & 51.20 & 36.41 & 29.73 & 39.75 & 49.22 & 36.49 & 32.88 & 38.89 & 42.95 \\
\midrule
\multirow{2}{*}{\makecell{LLM-based\\MMKGC}}
& MKGL & 32.86 & 26.54 & 35.57 & 44.76 & 29.11 & 24.30 & 31.60 & 38.28 & 27.14 & 18.68 & 30.39 & 43.87 \\
& K-ON & 38.10 & 30.13 & 42.77 & 53.59 & 36.64 & 30.05 & 38.72 & 48.26 & - & - & - & - \\
\midrule
Ours & RADD & \textbf{42.80} & \textbf{34.82} & \textbf{46.83} & \textbf{57.94} & \textbf{40.39} & \textbf{34.34} & \textbf{42.81} & \textbf{51.79} & \textbf{42.56} & \textbf{38.17} & \textbf{43.80} & \textbf{50.98} \\
\bottomrule
\end{tabular}
}
\caption{Filtered results (\%; higher is better). Best values are \textbf{bold}; the strongest published baselines are \underline{underlined}. Dashes denote unreported results.}
\label{tab:main}
\end{table*}

\subsection{Baselines}

We compare against 30 baselines in four groups.
\textbf{Unimodal KGC:} TransE~\cite{bordes2013transe}, TransD~\cite{ji2015transd}, DistMult~\cite{yang2015distmult}, ComplEx~\cite{trouillon2016complex}, RotatE~\cite{sun2019rotate}, PairRE~\cite{chao2021pairre}, GC-OTE~\cite{tang2020gcote}.
\textbf{Multimodal KGC:} IKRL~\cite{xie2017ikrl}, TBKGC~\cite{sergieh2018tbkgc}, TransAE~\cite{wang2019transae}, MMKRL~\cite{lu2022mmkrl}, RSME~\cite{wang2021rsme}, VBKGC~\cite{zhang2022qeb}, OTKGE~\cite{cao2022otkge}, MoSE~\cite{zhao2022mose}, IMF~\cite{li2023imf}, QEB~\cite{wang2023tivakg}, VISTA~\cite{lee2023vista}, AdaMF and AdaMF-MAT~\cite{zhang2024adamfmat}, and MoMoK~\cite{zhang2025momok} explore multimodal representation learning, modality interaction, and prediction fusion. SNAG~\cite{chen2025snag} uses Transformer-based interaction and modality-noise masking; MCKGC~\cite{gao2025mckgc} fuses representations across hyperbolic, hyperspherical, and Euclidean spaces. ReranKGC~\cite{gao2025rerankgc} combines an embedding retriever with a prompt-tuned KGC-CLIP reranker.
\textbf{Negative sampling:} KBGAN~\cite{cai2018kbgan}, MANS~\cite{zhang2023mans}, MMRNS~\cite{xu2022mmrns}, and DHNS~\cite{niu2025dhns} improve training-time sample quality.
\textbf{LLM-augmented:} MKGL~\cite{guo2024mkgl} adapts language models to a specialized KG language; K-ON~\cite{guo2025kon} uses multiple prediction heads for entity-level prediction.
K-ON, SNAG, and MCKGC results come from their original papers. RADD (retriever only) uses the frozen retriever without refinement. For Table~\ref{tab:directional}, we run AdaMF-MAT and MoMoK using the original authors' released code. These directional runs are separate from the published aggregate results in Table~\ref{tab:main}; their averages need not match those entries.

\begin{table*}[t]
\centering
\footnotesize
\setlength{\tabcolsep}{2.5pt}
\renewcommand{\arraystretch}{1.5}
\resizebox{\textwidth}{!}{%
\begin{tabular}{@{}l*{18}{c}@{}}
\toprule
\multirow{3}{*}{Model} & \multicolumn{6}{c}{DB15K} & \multicolumn{6}{c}{MKG-W} & \multicolumn{6}{c}{MKG-Y} \\
\cmidrule(lr){2-7}\cmidrule(lr){8-13}\cmidrule(lr){14-19}
& \multicolumn{3}{c}{Head} & \multicolumn{3}{c}{Tail}
& \multicolumn{3}{c}{Head} & \multicolumn{3}{c}{Tail}
& \multicolumn{3}{c}{Head} & \multicolumn{3}{c}{Tail} \\
\cmidrule(lr){2-4}\cmidrule(lr){5-7}\cmidrule(lr){8-10}\cmidrule(lr){11-13}\cmidrule(lr){14-16}\cmidrule(lr){17-19}
& MRR & H@1 & H@3 & MRR & H@1 & H@3
& MRR & H@1 & H@3 & MRR & H@1 & H@3
& MRR & H@1 & H@3 & MRR & H@1 & H@3 \\
\midrule
AdaMF-MAT & 23.94 & 16.67 & 27.59 & 47.94 & 35.51 & 56.14
& \underline{18.56} & 13.10 & \underline{20.57} & \underline{52.80} & 44.85 & \underline{56.79}
& \underline{39.48} & 34.40 & \underline{41.98} & \underline{38.75} & 34.81 & \underline{40.48} \\
MoMoK & \underline{26.23} & \underline{19.69} & \underline{28.94} & \underline{51.68} & \underline{43.18} & \underline{56.29}
& 18.50 & \underline{13.59} & 19.82 & 51.66 & \underline{45.95} & 54.14
& 37.79 & \underline{34.81} & 39.58 & 37.49 & \underline{35.19} & 38.42 \\
RADD & \textbf{29.31} & \textbf{22.76} & \textbf{32.00} & \textbf{56.29} & \textbf{46.88} & \textbf{61.65}
& \textbf{22.48} & \textbf{17.06} & \textbf{24.17} & \textbf{58.29} & \textbf{51.61} & \textbf{61.44}
& \textbf{43.28} & \textbf{38.57} & \textbf{44.57} & \textbf{41.84} & \textbf{37.78} & \textbf{43.03} \\
\bottomrule
\end{tabular}}
\caption{Head and tail prediction results (\%; higher is better) on three benchmarks. Baselines are reproduced using the original authors' released code. Best results are \textbf{bold}; the strongest reproduced baseline is \underline{underlined}.}
\label{tab:directional}
\end{table*}

\subsection{Research Questions}
\label{sec:rqs}

Our experiments address five questions: (RQ1) Does RADD improve strong baselines and its matched retriever? (RQ2) How does refinement perform across head and tail prediction? (RQ3) Which components drive gains? (RQ4) How sensitive is performance to key hyperparameters? (RQ5) How do candidate coverage and refinement behavior explain practical performance?

\subsection{Results and Analysis}
\label{sec:results}

\subsubsection{Main Results (RQ1)}

\textbf{Unimodal baselines.} As shown in Table~\ref{tab:main}, RADD exceeds GC-OTE/PairRE/RotatE, the strongest unimodal MRR baselines on DB15K/MKG-W/MKG-Y, by 10.95/5.99/7.61 percentage points (pp), respectively. It complements structural scores with visual/textual evidence and query-conditioned comparison of plausible entities.

\textbf{Multimodal baselines.} RADD surpasses AdaMF-MAT and improves over MoMoK by 3.23/4.50/4.65 MRR pp. Beyond representation fusion, it learns evidence-driven corrections relative to retrieval preferences. Matched-retriever gains of 1.64/2.02/2.18 pp reported in Table~\ref{tab:abl} and favorable comparisons with pointwise/listwise rerankers on DB15K (Appendix~\ref{app:db15k-reranking}) further support this refinement design.

\textbf{Strongest reported baselines.} RADD outperforms MCKGC on DB15K and SNAG on MKG-W/Y by 3.01/3.09/3.06 MRR pp and leads all Hits metrics. Gains in both Top-1 accuracy and broader ranking quality support combining retrieval-anchored refinement with coverage-aware full-entity ranking.

\subsubsection{Head vs.\ Tail Entity Prediction (RQ2)}
\label{sec:directional}

Table~\ref{tab:directional} shows that RADD improves every reported metric in both directions. On DB15K/MKG-W/MKG-Y, head MRR gains over the strongest reproduced baseline are 3.08/3.92/3.81 pp and tail gains are 4.61/5.49/3.09 pp (before rounding).

DB15K and MKG-W remain tail-favored. Higher tail candidate recall (91.61\%/83.06\%) than head recall (63.44\%/39.68\%) offers more opportunities for direct answer refinement, helping explain this asymmetry. Missed answers remain eligible, and gains persist despite lower head coverage.

MKG-Y is more balanced, with RADD reaching head/tail MRR of 43.28/41.84. Together, these gains support bidirectional refinement through a shared denoiser trained on both query types and conditioned on directed relations.

\subsubsection{Ablation Study (RQ3)}
\label{sec:ablation}

\begin{table*}[t]
\centering
\footnotesize
\setlength{\tabcolsep}{3.5pt}
\resizebox{\textwidth}{!}{%
\begin{tabular}{lcccccccccccc}
\toprule
Variant & \multicolumn{4}{c}{DB15K} & \multicolumn{4}{c}{MKG-W} & \multicolumn{4}{c}{MKG-Y} \\
\cmidrule(lr){2-5}\cmidrule(lr){6-9}\cmidrule(lr){10-13}
& MRR & H@1 & H@3 & H@10 & MRR & H@1 & H@3 & H@10 & MRR & H@1 & H@3 & H@10 \\
\midrule
\textbf{Full RADD} & \textbf{42.80} & \textbf{34.82} & \textbf{46.83} & \textbf{57.94} & \textbf{40.39} & \textbf{34.34} & \textbf{42.81} & \textbf{51.79} & \textbf{42.56} & \textbf{38.17} & \textbf{43.80} & \textbf{50.98} \\
w/o diffusion & 41.16 & 32.87 & 45.29 & 56.90 & 38.37 & 32.05 & 40.57 & 50.16 & 40.38 & 35.35 & 42.26 & 49.42 \\
w/o retrieval scoring & 42.41 & 34.56 & 46.04 & 57.41 & 39.19 & 33.12 & 41.93 & 50.08 & 41.65 & 37.23 & 43.02 & 49.77 \\
w/o coverage moderation & 41.24 & 33.46 & 44.94 & 56.21 & 38.91 & 32.97 & 41.21 & 49.94 & 41.16 & 36.67 & 42.60 & 49.55 \\
w/o multimodal evidence & 42.08 & 34.14 & 45.84 & 57.31 & 38.68 & 32.53 & 40.97 & 50.40 & 41.25 & 36.91 & 42.66 & 49.14 \\
\bottomrule
\end{tabular}}
\caption{Component and decision-rule ablations (\%; higher is better). Full RADD achieves the best result in every column (\textbf{bold}).}
\label{tab:abl}
\end{table*}

\paragraph{Retrieval and denoising provide complementary evidence.}
Table~\ref{tab:abl} holds retrieval fixed. Without diffusion, ranking uses $s_{\mathrm{kge}}$ alone and loses 1.64/2.02/2.18 MRR pp. Removing final retrieval scores but retaining candidates and $\pi_{\mathrm{ret}}$ loses 0.40/1.20/0.91 pp, supporting complementary global retrieval and local correction. Masking visual/textual evidence only within refinement loses 0.72/1.71/1.31 MRR pp, demonstrating its additional value. Appendix~\ref{app:ablation-operations} specifies the operations.

\paragraph{Coverage moderation strengthens ranking updates.}
Fixing $c_{\mathrm{cov}}=1$ disables coverage moderation and reduces MRR by 1.56/1.48/1.40 pp. This supports adapting local ranking updates to predicted candidate coverage, with the largest benefit on DB15K.

\subsubsection{Hyperparameter Sensitivity (RQ4)}
\label{sec:sensitivity}

\begin{figure*}[t]
\centering
\includegraphics[width=\textwidth]{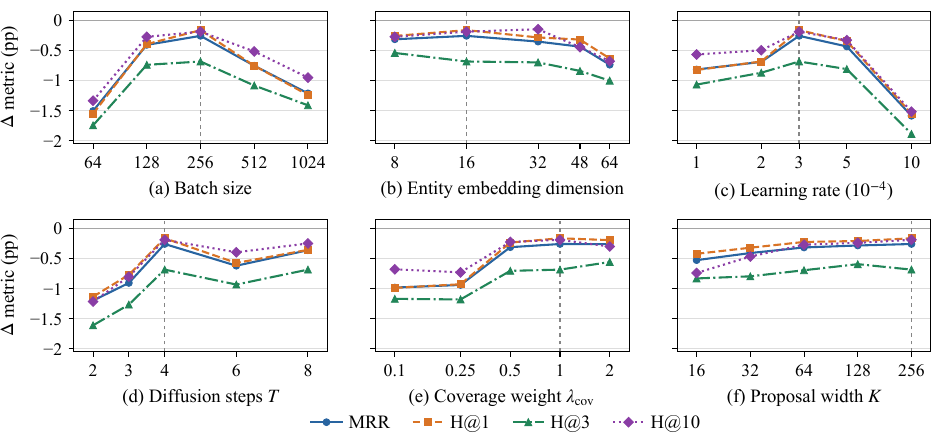}
\caption{DB15K sensitivity relative to RADD in Table~\ref{tab:main}. The vertical axis $\Delta$ metric is each sweep result minus the latest main-model reference, in percentage points (pp). Absolute reference MRR/H@1/H@3/H@10 is 42.80/34.82/46.83/57.94. Panels share a vertical scale; horizontal zero lines denote this reference and vertical dashed lines mark default parameter values. Original sweep observations are retained, so default points need not lie at zero.}
\label{fig:sensitivity}
\end{figure*}

On DB15K, we retain the original single-factor sweeps around batch size 256, entity embedding dimension 16, learning rate $3\times10^{-4}$, $T=4$, $\lambda_{\mathrm{cov}}=1$, and inference $K=256$. Each sweep varies one parameter while keeping the others fixed. Fig.~\ref{fig:sensitivity} expresses these unchanged observations relative to the latest full RADD result in Table~\ref{tab:main}, whose MRR/H@1/H@3/H@10 is 42.80/34.82/46.83/57.94\%.

The sweeps show limited sensitivity to entity embedding dimension and proposal width, with MRR ranges of 0.48 and 0.27 pp. Coverage weights from 0.5 to 2 yield an MRR range of only 0.06 pp; the full weight sweep spans 0.72 pp. Two to eight reverse steps yield MRR of 41.60 to 42.54, with four steps providing the strongest result. Batch size and learning rate span 1.25 and 1.33 pp, warranting more careful tuning.

\subsubsection{Retrieval Coverage and Refinement Behavior (RQ5)}
\label{sec:coverage_efficiency}

In the DB15K sensitivity study, increasing inference $K$ from 16 to 256 raises candidate recall from 54.91\% to 77.52\%, while MRR rises from 42.27 to 42.54 (+0.27 pp). From $K=64$ to 256, the additional MRR gain is only 0.06 pp despite an 11.15-pp recall increase, showing stable ranking performance across candidate budgets. Recall measures target inclusion in the expanded union, not a hard limit on full-entity prediction.

Query-level analyses show positive net correction: 921/359/207 Top-1 repairs versus 534/164/57 damaged predictions yield 387/195/150 additional correct answers on DB15K/MKG-W/MKG-Y. Cases in Appendix~\ref{app:casestudy} illustrate improved final rankings: RADD promotes the Tillamook County state answer from rank 26 to 1 and the Adam Lambert head-prediction answer from 12 to 1.

Refinement uses at most 288 candidates on DB15K and 96 on MKG-W/Y. Its complexity is $O(T(K_qd_{\mathrm{den}}^2+K_q^2d_{\mathrm{den}}))$, with quadratic rather than cubic reverse-kernel evaluation (Appendix~\ref{app:algorithm}). On DB15K, full RADD adds only 0.43\% to the parameter count of the frozen retriever (Appendix~\ref{app:db15k-reranking}).

\begingroup
\raggedbottom
\section{Conclusion}

To address retrieval ambiguity and incomplete candidate coverage in MMKGC, we propose RADD, which uses the query-specific retrieval distribution as both the diffusion endpoint and the reference for evidence-conditioned entity recovery. Full-state reverse propagation refines candidate preferences using multimodal evidence, while coverage-moderated density-ratio corrections preserve full-entity ranking without excluding candidate-external answers. Across three benchmarks, RADD outperforms the strongest reported baselines and improves matched-retriever MRR by 1.64, 2.02, and 2.18 percentage points. Directional results support effectiveness in both directions, while ablations demonstrate the value of multimodal evidence and coverage moderation.
\par\clearpage
\endgroup

\section{Limitations}

RADD preserves full-entity ranking while concentrating evidence-based refinement within a retrieved candidate pool. Under the current fixed proposal policy, entities outside this pool retain their retrieval scores. A natural direction is to make candidate expansion query-adaptive, using retrieval uncertainty and available multimodal evidence to decide where additional refinement is most useful. This could extend the reach of local evidence without uniformly increasing the candidate budget.

Extending this formulation from fixed benchmark graphs to larger, evolving knowledge graphs also opens new research opportunities. The current approach builds cross-fitted evidence from a fixed training graph and propagates probabilities over all candidate states. Incremental evidence updates and sparse candidate interactions are promising directions for maintaining retrieval and refinement as new facts arrive, while controlling training and inference costs.

\bibliography{radd_acl_submission}

\clearpage
\appendix
\raggedbottom
\nolinenumbers
\section{Implementation Details and Derivations}
\label{app:algorithm}

This appendix specifies candidate construction and the evidence network, derives the quadratic reverse-kernel evaluation, and characterizes the ranking correction. Algorithms~\ref{alg:train} and~\ref{alg:infer} summarize the training and inference procedures from Section~\ref{sec:approach}.

\begin{algorithm}[H]
\caption{RADD Cross-Fitted Training}
\label{alg:train}
\footnotesize
\begin{algorithmic}[1]
\Require Training triples partitioned into held-out folds $\mathcal T_f$ ($f=1,2,3$), schedule $\{\beta_t\}_{t=1}^T$, proposal width $K$, coverage weight $\lambda_{\mathrm{cov}}$, learning rate $\eta$
\Statex \textit{Stage 1: natural cross-fitted examples}
\For{each held-out fold $f$}
  \State Fit a fold-specific retriever and build structural evidence using $\mathcal T\setminus\mathcal T_f$
  \For{each $(h,r,t_e)\in\mathcal T_f$}
    \State Form tail pair $((h,r,?),t_e)$ and head pair $((?,r,t_e),h)$, each denoted $(q,y)$
    \State Cache $(q,y)$, fold-specific $s_{\mathrm{kge}}$, natural $\mathcal C_q$, $\pi_{\mathrm{ret}}$, descriptors, and $z_q=\mathbb I[y\in\mathcal C_q]$
  \EndFor
\EndFor
\Statex \textit{Stage 2: bidirectional denoising and coverage}
\State Initialize $\theta_{\mathrm{den}}$
\For{each training epoch}
\For{each mini-batch of cached queries}
  \State Predict $c_{\mathrm{cov}}$ and compute coverage BCE on all queries
  \For{each covered query ($z_q=1$)}
    \State Sample $t\sim\mathrm{Uniform}\{1,\ldots,T\}$ and $x_t\sim q_{\mathrm{fwd}}(\cdot\mid x_0=y,q,t)$
    \State Compute $\mathbf H_{q,t}$ and $\tilde{\mathbf x}(q,t,x_t)$
    \State Accumulate $-\log p_{\theta_{\mathrm{den}}}(x_0=y\mid x_t,q,t)$
  \EndFor
  \State Average the clean loss over covered queries (zero if none) and coverage BCE over all queries, as in Eqs.~\ref{eq:diff_loss}--\ref{eq:total_loss}
  \State Update $\theta_{\mathrm{den}}$ using $\nabla_{\theta_{\mathrm{den}}}\mathcal L$ and learning rate $\eta$; keep cache fixed
\EndFor
\EndFor
\State Fit evaluation retrieval $\theta_{\mathrm{ret}}$ on the complete training set; select the denoiser checkpoint without using test metrics
\end{algorithmic}
\end{algorithm}

\begin{algorithm}[!t]
\caption{RADD Inference with Reranking}
\label{alg:infer}
\footnotesize
\begin{algorithmic}[1]
\Require Query $q\in\{(h,r,?),(?,r,t_e)\}$, proposal width $K$, schedule $\{\beta_t\}_{t=1}^T$, frozen $\theta_{\mathrm{ret}}$, learned $\theta_{\mathrm{den}}$
\Ensure All entities ranked by $S_{\mathrm{dr}}$
\Statex \textit{Stage 1: global candidate retrieval}
\State Compute $s_{\mathrm{kge}}(e\mid q)$ for every $e \in \mathcal{E}$
\State $\mathcal{C}_K(q) \leftarrow \operatorname{TopK}_{e\in\mathcal{E}} s_{\mathrm{kge}}(e\mid q)$
\State Expand/deduplicate $\mathcal C_q$ via Eq.~\ref{eq:natural_candidates}; compute $\pi_{\mathrm{ret}}$ and candidate evidence
\State Predict $c_{\mathrm{cov}}$; initialize $\mu_T\leftarrow\pi_{\mathrm{ret}}$
\Statex \textit{Stage 2: full-state reverse propagation}
\For{$t=T,T-1,\ldots,1$}
  \State Compute $\mathbf H_{q,t}$
  \For{each current state $j\in\mathcal C_q$}
    \State Set $x_t=j$, $\mathbf x_t=\mathbf a_j$; construct $\tilde{\mathbf x}(q,t,x_t)$
    \State Evaluate $p_{\theta_{\mathrm{den}}}(x_0=\cdot\mid x_t=j,q,t)$ by Eq.~\ref{eq:reverse}
  \EndFor
  \State Form $R_t^q$ via Eq.~\ref{eq:efficient_reverse}; propagate $\mu_{t-1}$ by Eq.~\ref{eq:reverse_propagation}
\EndFor
\Statex \textit{Stage 3: coverage-moderated full-entity ranking}
\State $p_{\mathrm{diff}}\leftarrow\mu_0$; form $p_{\mathrm{OS}}$ via Eq.~\ref{eq:coverage_mixture}
\State Initialize $S_{\mathrm{dr}}\leftarrow s_{\mathrm{kge}}$ on $\mathcal E$
\For{each $e\in\mathcal C_q$}
  \State Add $\log p_{\mathrm{OS}}(e\mid q)-\log\pi_{\mathrm{ret}}(e\mid q)$ to its score
\EndFor
\State \Return all entities ranked by $S_{\mathrm{dr}}(e\mid q)$
\end{algorithmic}
\end{algorithm}

\subsection{Notation and Operator Conventions}
For a set, $|\cdot|$ denotes cardinality; $\cup$ and $\setminus$ denote union and set difference. Candidate vectors use a common entity ordering: $[\mathbf z]_e$ extracts the logit for entity $e$, and matrix subscripts index rows and columns. Superscript $\top$ denotes transpose, and $\mathbb R^{m\times n}$ denotes real matrices with $m$ rows and $n$ columns. All $\log$ and $\exp$ use the natural base; $\propto$ means proportionality before normalization. Distributions without entity arguments denote their probability vectors. Summation variables such as $e'$ and product indices such as $u$ are local dummy indices. The superscript $q$ in transition kernels marks query dependence. In Algorithm~\ref{alg:train}, $\sim$ denotes sampling, $\mathrm{Uniform}\{1,\ldots,T\}$ assigns probability $1/T$ to each step, BCE is binary cross-entropy, and $\nabla_{\theta_{\mathrm{den}}}\mathcal L$ is the gradient with respect to trainable parameters. The arrow $\leftarrow$ denotes assignment; query arguments are omitted when fixed.

\subsection{Candidate Evidence and Denoiser~Parameterization}

\paragraph{Retrieval configuration.}
\label{app:retrieval-config}
The reported ComplEx-based retrieval backend aggregates three independently trained multimodal retrievers after query-wise score normalization. Their collective parameters are denoted by $\theta_{\mathrm{ret}}$, and $s_{\mathrm{kge}}$ denotes the deployed score. This configuration is fixed across RADD, RADD (retriever only), and the same-retriever controls, isolating the contribution of refinement.

Two score conventions serve distinct roles. The deployed $s_{\mathrm{kge}}$ selects candidates and enters the final score in Eq.~\ref{eq:diff_rerank}. Its query-wise standardized version, $b_{q,e}=\mathrm Z_{\mathcal E}(s_{\mathrm{kge}}(e\mid q))$, defines $\pi_{\mathrm{ret}}$ in Eq.~\ref{eq:retriever_reference}: the mean and stabilized standard deviation are computed over all entities, whereas the softmax denominator includes only $\mathcal C_q$. The correction is added to $s_{\mathrm{kge}}$, not to $b_{q,e}$ or $\log\pi_{\mathrm{ret}}$. This distinction preserves the deployed retrieval scale in all matched comparisons.

The natural candidate union is
\begin{equation}
\begin{aligned}
\mathcal C_q={}&\mathcal C_K(q)\cup\mathcal N_v^8(e^*)\\
&\cup\mathcal N_t^8(e^*)\cup\mathcal U^{16}(q),
\end{aligned}
\label{eq:natural_candidates}
\end{equation}
where $e^*=\arg\max_{e\in\mathcal E}s_{\mathrm{kge}}(e\mid q)$ is the highest-scoring entity returned by $\arg\max$. The sets $\mathcal N_v^8(e^*)$ and $\mathcal N_t^8(e^*)$ contain its eight nearest visual and textual neighbors; their subscripts identify modalities, not diffusion time. The set $\mathcal U^{16}(q)$ supplies 16 deterministic proposals from the observed entity and relation. Superscripts 8 and 16 are proposal counts, not powers. Deduplication gives $K_q\leq K+32$. Coverage is measured on this union, not only the top-$K$ proposals.

\paragraph{Coverage prediction and candidate recall.}
The label $z_q$ records whether the designated answer $y$ belongs to the expanded, deduplicated $\mathcal C_q$. It supervises the learned inclusion probability $c_{\mathrm{cov}}(q)$, not the probability that the highest-ranked candidate is correct. At inference, only the predicted probability enters Eq.~\ref{eq:coverage_mixture}; the answer and its membership label are unavailable. In contrast, reported candidate recall is an empirical evaluation statistic. For a set $\mathcal Q$ of query--target pairs, it is
\[
\operatorname{Recall}(\mathcal Q)=\frac{1}{|\mathcal Q|}
\sum_{(q,y)\in\mathcal Q}\mathbb I[y\in\mathcal C_q].
\]
We compute recall separately for head and tail queries and micro-average over both directions for aggregate results. It is neither an average of predicted $c_{\mathrm{cov}}$ values nor a ceiling on full-entity ranking accuracy: entities outside $\mathcal C_q$ retain their retrieval scores.

\paragraph{Ablation operations.}
\label{app:ablation-operations}
The four controls in Table~\ref{tab:abl} isolate different parts of refinement while retaining the same retrieval backend and filtered evaluation.
\emph{w/o diffusion} disables the entire refinement correction and ranks directly by $s_{\mathrm{kge}}$.
\emph{w/o retrieval scoring} removes only the additive $s_{\mathrm{kge}}$ term from Eq.~\ref{eq:diff_rerank}, retaining candidate construction, $\pi_{\mathrm{ret}}$, and diffusion. Its remaining score is $\ell_{\mathrm{diff}}$, which is zero outside $\mathcal C_q$ under that equation; this control is not a retriever-independent model or a ranking by $p_{\mathrm{diff}}$ alone.
\emph{w/o coverage moderation} fixes $c_{\mathrm{cov}}(q)=1$ in Eq.~\ref{eq:coverage_mixture}, giving $p_{\mathrm{OS}}=p_{\mathrm{diff}}$ and removing coverage-dependent shrinkage toward $\pi_{\mathrm{ret}}$.
\emph{w/o multimodal evidence} masks visual/textual evidence within refinement while retaining structural evidence and the multimodal retriever. It tests refinement-side evidence, not the removal of all modalities from RADD.

\paragraph{Candidate evidence.}

The 24-dimensional $\mathbf f_{q,e}$ contains a 16-channel block describing visual/textual similarity, one-/two-hop propagation, relation-role signatures, answer prototypes, local relation neighborhoods, and entity degrees. Eight further channels summarize relation-conditioned visual/textual prototypes, modality availability, and relation--modality coherence. Graph-derived statistics, including answer prototypes, exclude the held-out fold when constructing cross-fitted training examples.

Multi-layer perceptrons (MLPs) encode each candidate and the query:
\begin{equation}
\begin{aligned}
\mathbf h_e^{(0)}&=\operatorname{MLP}_e([\mathbf a_e;\mathbf f_{q,e}]),\\
\mathbf u_{q,t}&=\operatorname{MLP}_q([\mathbf c_q;\mathbf r;\phi(t)]).
\end{aligned}
\label{eq:evidence_encodings}
\end{equation}
The shared candidate encoder $\operatorname{MLP}_e$ produces initial representation $\mathbf h_e^{(0)}$; $\operatorname{MLP}_q$ produces query/timestep context $\mathbf u_{q,t}$. Both have width $d_{\mathrm{den}}$, and superscript $(0)$ marks the pre-interaction representation. We then modulate candidates:
\begin{equation}
\begin{aligned}
\relax[\boldsymbol\gamma_{q,t};\boldsymbol\xi_{q,t}]&=W_F\mathbf u_{q,t},\\
\widetilde{\mathbf h}_e&=\operatorname{LN}\bigl(\mathbf h_e^{(0)}\odot
[1+0.1\tanh\boldsymbol\gamma_{q,t}]\\
&\hspace{5em}+\boldsymbol\xi_{q,t}\bigr).
\end{aligned}
\label{eq:evidence_modulation}
\end{equation}
Here the learned projection $W_F\in\mathbb R^{2d_{\mathrm{den}}\times d_{\mathrm{den}}}$ produces scale/shift vectors $\boldsymbol\gamma_{q,t},\boldsymbol\xi_{q,t}\in\mathbb R^{d_{\mathrm{den}}}$. The element-wise hyperbolic tangent $\tanh$ bounds the scale adjustment by 0.1; scalar 1 is broadcast across channels. The operators $\odot$ and $\operatorname{LN}$ denote element-wise multiplication and layer normalization, yielding modulated representation $\widetilde{\mathbf h}_e$. Multi-head self-attention and a feed-forward block produce contextual vectors $\mathbf h_e\in\mathbb R^{d_{\mathrm{den}}}$, stacked as rows $\mathbf h_e^\top$ in $\mathbf H_{q,t}$. Dependence of $\mathbf h_e^{(0)}$ on $q$, and of $\widetilde{\mathbf h}_e,\mathbf h_e$ on $(q,t)$, is suppressed for readability. The state-conditioned output of $f_{\theta_{\mathrm{den}}}$ is
\begin{equation}
\begin{aligned}
\relax[\mathbf z(q,t,x_t)]_e={}&\frac{(W_K\mathbf h_e)^\top
[\mathbf u_{q,t}+\operatorname{MLP}_s(\mathbf x_t)]}{\sqrt{d_{\mathrm{den}}}}\\
&+\operatorname{MLP}_b([\mathbf h_e;\mathbf u_{q,t}]).
\end{aligned}
\label{eq:evidence_logits}
\end{equation}
The learned projection $W_K\in\mathbb R^{d_{\mathrm{den}}\times d_{\mathrm{den}}}$ maps candidates to keys; its subscript labels this role, not proposal width $K$. The network $\operatorname{MLP}_s$ maps the current-state embedding to the hidden space, and $\operatorname{MLP}_b$ returns a scalar candidate-query bias. Thus Eq.~\ref{eq:evidence_logits} combines a dot product scaled by $\sqrt{d_{\mathrm{den}}}$ with an additive evidence term. Subscripts on the MLPs identify their roles; their parameters belong to $\theta_{\mathrm{den}}$.

\subsection{Reverse-Kernel Derivation and Complexity}

For a fixed query, let $\pi$ be the column vector of probabilities $\pi_e=\pi_{\mathrm{ret}}(e\mid q)$. Define the $K_q\times K_q$ forward matrix by $[Q_t]_{ij}=Q_t^q(j\mid i)$, with source states indexing rows. Then $Q_t=\alpha_tI+(1-\alpha_t)\mathbf1\pi^\top$, where $I$ is the identity matrix and $\mathbf1\in\mathbb R^{K_q}$ is the all-ones column vector; $\mathbf1\pi^\top$ repeats the retrieval distribution in every row. Since $(\mathbf1\pi^\top)^2=\mathbf1\pi^\top$, the cumulative transition $\bar Q_t=Q_1\cdots Q_t$ equals $\bar\alpha_tI+(1-\bar\alpha_t)\mathbf1\pi^\top$, proving Eq.~\ref{eq:forward}; $\bar Q_0=I$.

Bayes' rule gives the reverse posterior $q_{\mathrm{rev}}$, the probability of $x_{t-1}=i$ given $x_t=j$, hypothesized $x_0=e$, and query $q$:
\begin{equation}
q_{\mathrm{rev}}(i\mid j,e,q,t)=
\frac{\bar Q_{t-1}^q(i\mid e)Q_t^q(j\mid i)}{\bar Q_t^q(j\mid e)},
\label{eq:reverse_posterior}
\end{equation}
where $i,j,e\in\mathcal C_q$ denote previous, current, and hypothesized clean identities. Both the posterior and its mixing distribution in Eq.~\ref{eq:learned_reverse} are normalized, so $\sum_iR_t^q(i\mid j)=1$. Unlike the forward matrix, $[R_t^q]_{ij}=R_t^q(i\mid j)$ uses destination rows and source columns; hence Eq.~\ref{eq:reverse_propagation} is $\mu_{t-1}=R_t^q\mu_t$. The denominator in Eq.~\ref{eq:reverse_posterior} depends on $e$ and must be applied before mixing clean identities.

The mixture need not be computed as a cubic three-index sum. Define
\begin{equation}
\begin{aligned}
A&=\bar\alpha_{t-1},\qquad B=1-A,\\
W_{e,j}&=\frac{p_{\theta_{\mathrm{den}}}(x_0=e\mid x_t=j,q,t)}{\bar Q_t^q(j\mid e)},\\
S_j&=\sum_{e\in\mathcal C_q}W_{e,j}.
\end{aligned}
\label{eq:reverse_weights}
\end{equation}
Here $A$ and $B$ are the cumulative retention and replacement weights at $t-1$. The quantity $W_{e,j}$ divides the predicted clean-entity probability by its forward marginal, and $S_j$ is its column sum over clean entities. These are computed scalars, not learned weights; dependence on $(q,t)$ is suppressed.
Substituting $\bar Q_{t-1}^q(i\mid e)=A\mathbb I[i=e]+B\pi_i$ into Eq.~\ref{eq:learned_reverse} gives
\begin{equation}
R_t^q(i\mid j)=Q_t^q(j\mid i)[A W_{i,j}+B\pi_i S_j],
\label{eq:efficient_reverse}
\end{equation}
where $\pi_i=\pi_{\mathrm{ret}}(i\mid q)$. This requires $O(K_q^2)$ operations per step; $O(\cdot)$ denotes asymptotic cost up to constant factors. Including attention and hidden projections, refinement costs $O(T(K_qd_{\mathrm{den}}^2+K_q^2d_{\mathrm{den}}))$ time and $O(K_q^2+K_qd_{\mathrm{den}})$ inference working memory, in addition to retrieval. Stabilized probability or log-space arithmetic is needed for small reference probabilities.

\subsection{Properties of the Ranking Correction}

\paragraph{Identity case.}
At $t=1$, Eq.~\ref{eq:reverse_posterior} reduces to $\mathbb I[i=e]$, so $R_1^q(i\mid j)=p_{\theta_{\mathrm{den}}}(x_0=i\mid x_1=j,q,1)$. If the evidence logits are constant across clean candidates for every current state at this step, each column of $R_1^q$ equals $\pi$. Consequently, $\mu_0=\pi$ for any normalized $\mu_1$, yielding the retrieval fallback in Eq.~\ref{eq:diff_rerank}.

\makeatletter
\setlength{\@dblfptop}{0pt}
\setlength{\@dblfpsep}{1em}
\makeatother
\begin{table*}[t]
\centering
\small
\setlength{\tabcolsep}{3pt}
\renewcommand{\arraystretch}{1.2}
\resizebox{\textwidth}{!}{%
\begin{tabular}{@{}l*{12}{r}@{}}
\toprule
\multirow{2}{*}{Model} & \multicolumn{4}{c}{FB15K-237} & \multicolumn{4}{c}{NELL995} & \multicolumn{4}{c}{YAGO3-10}\\
\cmidrule(lr){2-5}\cmidrule(lr){6-9}\cmidrule(lr){10-13}
& MRR & H@1 & H@3 & H@10 & MRR & H@1 & H@3 & H@10 & MRR & H@1 & H@3 & H@10\\
\midrule
TransE & 33.10 & 23.40 & 36.90 & 52.60 & 23.90 & 7.90 & 34.80 & 50.80 & 32.40 & 22.10 & 37.40 & 52.40\\
DistMult & 28.40 & 22.30 & 31.00 & 44.70 & 34.80 & 26.80 & 38.50 & 50.00 & 35.70 & 25.20 & 40.70 & 56.80\\
ComplEx & 30.80 & 22.00 & 33.70 & 48.60 & \underline{36.90} & \underline{28.90} & \underline{40.50} & 51.80 & 38.30 & 27.70 & 44.00 & 58.90\\
RotatE & \underline{33.60} & \underline{24.00} & \underline{37.40} & \underline{53.00} & 35.80 & 26.90 & 40.40 & \textbf{\underline{52.60}} & \underline{41.90} & \underline{32.10} & \underline{47.50} & \underline{60.70}\\
\midrule
RADD (retriever only) & 32.67 & 23.30 & 36.06 & 51.57 & 39.29 & 34.03 & 42.43 & 47.58 & 47.07 & 37.58 & 51.98 & 65.55\\
RADD & \textbf{34.70} & \textbf{25.23} & \textbf{38.32} & \textbf{53.80} & \textbf{41.08} & \textbf{35.40} & \textbf{44.63} & 50.63 & \textbf{52.19} & \textbf{44.45} & \textbf{56.33} & \textbf{66.54}\\
\midrule
$\Delta$ (full $-$ retriever) & +2.03 & +1.93 & +2.26 & +2.24 & +1.78 & +1.37 & +2.19 & +3.04 & +5.12 & +6.87 & +4.35 & +0.99\\
\bottomrule
\end{tabular}}
\caption{Structure-only knowledge graph completion under full-entity filtered evaluation (\%; higher is better), averaged over head and tail prediction. MRR denotes mean reciprocal rank; H@$k$ denotes Hits@$k$. Best results are \textbf{bold}; the strongest literature baseline is \underline{underlined}. $\Delta$ is full RADD minus its retriever in percentage points, computed before rounding. RADD improves over its retriever in all four metrics on every dataset.}
\label{tab:structure-only}
\end{table*}

\paragraph{Nonlinear moderation and its bound.}
For $e\in\mathcal C_q$, combining Eqs.~\ref{eq:coverage_mixture} and~\ref{eq:diff_rerank} gives
\begin{equation*}
\begin{aligned}
\ell_{\mathrm{diff}}(e\mid q)=\log\!\Bigl[&1-c_{\mathrm{cov}}(q)\\
&+c_{\mathrm{cov}}(q)\frac{p_{\mathrm{diff}}(e\mid q)}{\pi_{\mathrm{ret}}(e\mid q)}\Bigr].
\end{aligned}
\end{equation*}
Thus coverage is not a linear weight on the log ratio. Since $p_{\mathrm{diff}}/\pi_{\mathrm{ret}}\geq0$, the correction satisfies $\ell_{\mathrm{diff}}\geq\log(1-c_{\mathrm{cov}})$ for $c_{\mathrm{cov}}<1$. This limits downward score changes but does not guarantee preservation of a correct Top-1 prediction.

\paragraph{Score-scale dependence.}
Standardizing scores to construct $\pi_{\mathrm{ret}}$ does not make Eq.~\ref{eq:diff_rerank} invariant to rescaling the deployed $s_{\mathrm{kge}}$: the reference can remain unchanged while the relative magnitude of the additive correction changes. The retriever's score aggregation and normalization are therefore part of the fixed scoring convention.

\ifdefined\linenumbers\linenumbers\fi

\begin{table*}[!t]
\centering
\small
\setlength{\tabcolsep}{6pt}
\begin{tabular}{llrrrrrr}
\toprule
Dataset & Direction & Queries & \makecell{RADD\\(retriever only)} & RADD & Repaired & Damaged & Net\\
\midrule
\multirow{2}{*}{DB15K} & Tail & 9,902 & 4,445 & \textbf{4,642} & 514 & 317 & \textbf{197}\\
& Head & 9,902 & 2,064 & \textbf{2,254} & 407 & 217 & \textbf{190}\\
\midrule
\multirow{2}{*}{MKG-W} & Tail & 4,274 & 2,084 & \textbf{2,206} & 180 & 58 & \textbf{122}\\
& Head & 4,274 & 656 & \textbf{729} & 179 & 106 & \textbf{73}\\
\midrule
\multirow{2}{*}{MKG-Y} & Tail & 2,663 & 938 & \textbf{1,006} & 104 & 36 & \textbf{68}\\
& Head & 2,663 & 945 & \textbf{1,027} & 103 & 21 & \textbf{82}\\
\bottomrule
\end{tabular}
\caption{Top-1 (rank-1) correct-prediction counts. RADD denotes the full model. Repaired/Damaged count incorrect-to-correct/correct-to-incorrect changes relative to retrieval alone; Net is their difference. Bold highlights full-model counts and positive net gains: RADD improves top-ranked accuracy on every dataset in both directions.}
\label{tab:radd-top1-full}
\end{table*}

\section{Generalization to Structure-Only KGC}
\label{app:structure-only}

\paragraph{Experimental scope.}
We further evaluate whether retrieval-anchored diffusion remains effective without multimodal inputs on FB15K-237, NELL995, and YAGO3-10. Both retrieval and refinement use structural information only, without images or text. Table~\ref{tab:structure-only} compares full RADD with its retriever-only counterpart and four representative KGE methods: TransE, DistMult, ComplEx, and RotatE. Evaluation ranks all entities with other known correct answers filtered out, averages head and tail results, and reports metrics as percentages.

\paragraph{Baseline sources.}
FB15K-237 and NELL995 baseline results follow Table~5 of \citet{sun2023sfca}. For YAGO3-10, DistMult and ComplEx follow Table~2 of \citet{balkir2019pairwise}, while TransE and RotatE follow the 32-dimensional results in Table~3 of \citet{wang2023greenkgc}. These literature results retain their original training configurations; the matched-retriever comparison isolates the added refinement within RADD.

\paragraph{Consistent gains over retrieval.}
RADD improves all four metrics on all three datasets, demonstrating the value of diffusion-based refinement even without complementary modalities. MRR gains are 2.03, 1.78, and 5.12 percentage points (pp) on FB15K-237, NELL995, and YAGO3-10, corresponding to relative improvements of 6.23\%, 4.54\%, and 10.87\%, respectively. On YAGO3-10, the 6.87-pp H@1 gain, compared with 0.99 pp in H@10, further indicates that refinement is particularly effective at improving top-ranked~predictions.

\paragraph{Comparison with KGE baselines.}
RADD achieves the best results in 11 of the 12 dataset and metric combinations in Table~\ref{tab:structure-only}. It exceeds the best listed KGE baseline in MRR by 1.10 pp on FB15K-237, 4.18 pp on NELL995, and 10.29 pp on YAGO3-10. On FB15K-237, retrieval alone trails RotatE in MRR, whereas full RADD surpasses it, showing that the final advantage is not simply inherited from a stronger retriever. On NELL995, RADD leads in MRR, H@1, and H@3; its H@10 of 50.63 remains below RotatE's 52.60, while improving over its own retriever by 3.04 pp. This pattern highlights stronger top-ranked accuracy alongside improvements across ranking cutoffs over retrieval alone.

\paragraph{Implications for the framework.}
These results show that RADD's effectiveness extends beyond multimodal fusion. Structural entity and relation evidence can also guide the revision of retrieval preferences through the same retrieval-anchored recovery formulation. The consistent gains over retrieval alone support diffusion-based refinement as a useful complement to global KGE scoring, while the multimodal experiments in the main text demonstrate its ability to incorporate additional visual and textual evidence.

\begin{table*}[!t]
\centering
\small
\renewcommand{\arraystretch}{1.2}
\setlength{\tabcolsep}{4pt}
\begin{tabular}{@{}lp{0.8cm}p{6.05cm}p{4.1cm}rr@{}}
\toprule
Dataset & Dir. & Query & Correct answer & Ret. & Full\\
\midrule
\multirow{2}{*}{DB15K} & Tail & (Tillamook County, Oregon, state, ?) & Oregon & 26 & \textbf{1}\\
& Head & (?, birthPlace, Tokyo) & Minami Takayama & 6 & \textbf{1}\\
\midrule
\multirow{2}{*}{MKG-W} & Tail & (Yayoi Kusama, P135, ?) & pop art & 2134 & \textbf{1}\\
& Head & (?, P175, Melvins) & Bullhead (album) & 6365 & \textbf{1}\\
\midrule
\multirow{2}{*}[-5pt]{MKG-Y} & Tail & (Cry, the Beloved Country (1995 film), \mbox{isLocatedIn}, ?) & South Africa & 28 & \textbf{1}\\
& Head & (?, created, For Your Entertainment (song)) & Adam Lambert & 12 & \textbf{1}\\
\bottomrule
\end{tabular}
\caption{Selected cases, with one head and one tail query per dataset. Dir. denotes prediction direction; Ret. and Full denote RADD (retriever only) and full RADD. Entries are filtered correct-answer ranks (lower is better); RADD promotes every selected answer to rank~1. Merged dataset cells group both directions, and P135/P175 are MKG-W relation identifiers. Entity-name underscores are replaced by spaces; head queries retain the original triple orientation. Table~\ref{tab:radd-top1-full} reports all-query repairs and damage beyond these selected successes.}
\label{tab:radd-cases}
\end{table*}

\section{Supplementary Query-Level Analyses}
\label{app:casestudy}

Tables~\ref{tab:radd-top1-full} and~\ref{tab:radd-cases} supplement Section~\ref{sec:coverage_efficiency} with directional counts and individual predictions, comparing RADD (retriever only) with full RADD. Entity embedding dimensions and inference proposal widths follow the main experiments.

\paragraph{Prediction transitions.}
A \textit{repair} changes an incorrect retriever Top-1 prediction into a correct full-model prediction; \textit{damage} is the converse. Their difference is the net increase in correct predictions. Table~\ref{tab:radd-top1-full} evaluates all test queries, including candidate misses. Across all three datasets, repairs outnumber damage for both query directions, yielding tail/head net gains of 197/190 on DB15K, 122/73 on MKG-W, and 68/82 on MKG-Y. Together, these gains increase average H@1 by 1.95/2.28/2.82 pp, respectively, showing that refinement improves top-ranked accuracy despite occasional harmful updates.

\paragraph{Interpreting individual corrections.}
Table~\ref{tab:radd-cases} illustrates how RADD improves final rankings across diverse relations. The Tillamook County state answer rises from rank 26 to 1, and the Adam Lambert head-prediction answer from 12 to 1. RADD also promotes the MKG-W answers from ranks 2,134 and 6,365 to 1, showing that expanded candidates can support corrections beyond the initial top-$K$ list. These large gains concern answers inside the expanded pool: a candidate-external answer keeps its score, and its rank can change by at most $K_q$ (at most 96 on MKG-W) under fixed filtering and tie handling. The cases illustrate successful refinement, while the all-query counts in Table~\ref{tab:radd-top1-full} establish its net benefit.

\section{Same-Retriever Reranking Controls}
\label{app:db15k-reranking}

We compare pointwise MLP, self-attention listwise, and RADD refinement on DB15K with the retrieval backend held fixed. The retriever-only and full-RADD rows in Table~\ref{tab:db15k-reranking} match the corresponding DB15K results in Table~\ref{tab:abl}.

\begin{table}[H]
\centering
\footnotesize
\setlength{\tabcolsep}{3pt}
\renewcommand{\arraystretch}{1.15}
\resizebox{\columnwidth}{!}{%
\begin{tabular}{lrrrr}
\toprule
Method & MRR & H@1 & H@3 & H@10\\
\midrule
RADD (retriever only) & 41.16 & 32.87 & 45.29 & 56.90\\
Pointwise MLP & 41.43 & 33.45 & 45.11 & 56.71\\
Self-attention listwise & 41.12 & 33.05 & 44.83 & 56.64\\
RADD & \textbf{42.80} & \textbf{34.82} & \textbf{46.83} & \textbf{57.94}\\
\bottomrule
\end{tabular}}
\caption{DB15K reranking results (\%; higher is better) with the same retriever. Best results are \textbf{bold}.}
\label{tab:db15k-reranking}
\end{table}

\paragraph{Effectiveness beyond generic reranking.}
RADD surpasses pointwise MLP and self-attention listwise reranking by 1.37 and 1.69 MRR pp, respectively, and leads all four metrics. Pointwise MLP yields a modest 0.27-pp MRR gain over retrieval alone, compared with 1.64 pp for RADD. These comparisons support RADD's refinement design over the tested non-diffusion alternatives with the same retrieval backbone.

\paragraph{Parameter overhead.}
Total parameters increase from 93,803,646 for retrieval alone to 94,208,960 for RADD, an addition of only 405,314 parameters (0.43\%). The listwise control has 94,185,280 parameters, only 23,680~fewer.

\end{document}